\pdfoutput=1

\documentclass[11pt,a4paper]{article}

\usepackage[hyperref]{ACL2023}

\usepackage{times}
\usepackage{latexsym}
\usepackage{tabularx}

\usepackage{graphicx}
\graphicspath{ {./images/} }

\usepackage[T1]{fontenc}

\usepackage[utf8]{inputenc}

\usepackage{microtype}

\usepackage{inconsolata}

\title{FlairNLP at SemEval-2023 Task 6b: Extraction of Legal Named Entities from Legal Texts using Contextual String Embeddings}

\author{Vinay N Ramesh \qquad Rohan Eswara
\\Univeristy of Colorado, Boulder \\
\tt\{vina2391, roes6117
\}@.colorado.edu}

\begin{document}
\maketitle
\begin{abstract}
Indian court legal texts and processes are essential towards the integrity of the judicial system and towards maintaining the social and political order of the nation. Due to the increase in number of pending court cases, there is an urgent need to develop tools to automate many of the legal processes with the knowledge of artificial intelligence. In this paper, we employ knowledge extraction techniques, specially the named entity extraction of legal entities within court case judgements. We evaluate several state of the art architectures in the realm of sequence labeling using models trained on a curated dataset of legal texts. We observe that a Bi-LSTM model trained on Flair Embeddings achieves the best results, and we also publish the BIO formatted dataset as part of this paper.
\end{abstract}

\section{Introduction}
The Legal Entity Extraction task \cite{kalamkar2022named} aims at developing a tool for the identification of named entities within Indian legal texts. Much of the Indian legal texts, such as court judgements are in English, however they assume a very unique format. This unstructured nature of Indian court judgements leads to a difficulty in parsing using simpler techniques such as regular expressions. Moreover, the entities which we are interested to extract are unique to the domain and already existing baseline models prove to be ineffective.
\\ \\
Techniques in NLP has made tremendous leaps in the last decade. While in the past, it would struggle to classify the sentiment of a sentence, the models today can classify text and generate sentences with almost no context \cite{topal2021exploring}. Many newer language models are trained on a general domain, but further fine-tuned to be used for a specific domain (e.g., science) \cite{jeong2022scideberta}. Indeed, these methods are achieving state-of-the-art results on Named Entity Recognition, Dependency Parsing and Relation Classification \cite{zhou2016attention} tasks.
\\ \\
In this paper, we propose training a deep neural language model using a labeled legal dataset for the task of Named Entity Recognition. We model a Bi-LSTM layer for token vectorization followed by a CRF layer for sequence labeling. To account for information from contexts, we use the Flair embeddings \cite{akbik-etal-2019-flair}, which is currently the state-of-the-art in sequence labeling tasks. Moreover, we curate the dataset used for training in the IOB format \cite{jiang2016evaluating} and release the dataset to the community.
\\
Besides the description discussed, we make the following observations from our experiments
\begin{itemize}
  \item Contextual string embeddings provide context to the sequence labeling tasks, improving the accuracy of identification of custom named entities.
  \item Bi-LSTM layer uses the context in both forward and backward direction to generate context vector for individual tokens
  \item The CRF layer uses these token probabilities to obtain the best path vector of sequence labels.
\end{itemize}

We also make the code available on this repository\footnote{\url{https://github.com/VinayNR/legaleval-2023}}.

\section{Background}
Named Entity Recognition (NER) \cite{nadeau2007survey} is an important natural language task which is used in Question Answering, Information Retrieval, Co-reference Resolution. Identification of named entities also paves way for word sense disambiguation and summarization tasks \cite{aliwy2021nerws}.
\\
Legal NER has been a topic of interest in the research community. \cite{dozier2010named} introduces NER on legal text and entity linking and resolution of those named entities. They categorize US legal texts into 5 classes - judges, attorneys, companies, courts and jurisdictions.
In the context of Indian legal system, \cite{Kalamkar2022CorpusFA} introduces structuring court judgements that are segmented into topical and coherent parts. They show the application of rhetorical roles to improve performance on legal summarization and judgement prediction.
\\
\cite{paul2022lesicin} proposes using a graph-based model for the task of legal statute identification. They enhance their learning by using the citation networks of legal documents along with textual data. In the space of court judgement predictions, \cite{malik-etal-2021-ildc} establishes the baseline of 78 percent accuracy.
\\
\cite{chalkidis2020legal} introduces LegalBERT which is a trained BERT model on legal corpus for specific downstream tasks.
\\ \\
We build on the existing knowledge of employing pre-trained models on a specific domain, along with contextual string embeddings to train a Bi-LSTM CRF model. In the domain of legal NER, we match the state-of-the-art results seen earlier.

\section{Model Architecture}
We introduce a contextual string embedding based deep neural architecture for the task of legal named entity recognition. Unlike many other language models \cite{devlin2018bert} trained on large corpus of text, we employ a character based language model. These contextual string embeddings allows us to pre-train on large, unlabeled corpus as well as learn different embeddings for the same words depending on the context.
Figure 1. explains the architecture of the model. Each input token Xi is passed through an embedding layer to get a vector representation. This is then provided as input to a Bi-LSTM layer which learns the contextual information of the words in a sentence. The CRF layer is then trained to learn the best path sequence from the output of the LSTM layer.
\\
\subsection{Problem Statement}
Formally introducing the problem, we have a set of tokens \begin{quote}
X = {x1, x2, ..., xn}
\end{quote} for which we need to identify spans of entities that are predefined. As per the task, we have 14 classes of entities to categorize - 
COURT, PETITIONER, RESPONDENT, JUDGE, LAWYER, DATE, ORG, GPE, STATUTE, PROVISION, PRECEDENT, CASENUMBER, WITNESS, OTHERPERSON
\\
We use the IOB formatted dataset to train, therefore the number of classes is effectively 29. We train a sequence labeling model to identify the named entity for a span of tokens and minimize the Viterbi Loss.

\subsection{Data Preparation}
\begin{figure}
\centering
\includegraphics[scale=0.4]{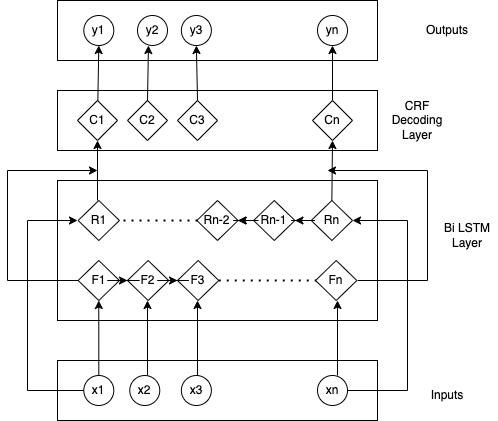}
\caption{\label{fig:The-caption}Model Architecture}
\end{figure}
The dataset consists of 11970 samples found in the Preamble and the Judgement where each sample is labeled for named entities. The dataset also has an equal distribution of classes to avoid problems concerning Imbalanced Classification \cite{kaur2019systematic}. Figure 2. and Figure 3. illustrates the class distribution in our training and validation dataset respectively. For training, we parse each of the samples and convert it to an IOB format and each token of a sample is on a new line identified by its corresponding tag. We remove stop words from each of the sentences and also purge all white-space characters.

\subsection{Mathematical Formulation}

\subsubsection{Bi-LSTM networks}
LSTMs are variants of Recurrent Neural Networks that have the ability to learn long-term dependencies in sequential data. The LSTM units contains special gates to control the flow of information into and out of these LSTM units, which are eventually used to form the LSTM network. 
Two networks stacked form the bidirectional LSTM, which learns contexts from both directions. This output is fed to the following CRF layer to predict the label sequence. The equations to update an LSTM unit or cell at each time step \textit{$t$}\hspace{5px}is given below :

\begin{equation}
  i_{t} = \sigma(W_{i} [x_{t} , h_{t{-}1}] + b_{i}),
\end{equation}
\begin{equation}
  f_{t} = \sigma(W_{f} [x_{t} , h_{t{-}1}] + b_{f}), 
\end{equation}
\begin{equation}
  o_{t} = \sigma(W_{o}[x_{t} , h_{t{-}1}] + b_{o}),
\end{equation}
\begin{equation}
  c˜ = tanh(W_{c}[x_{t} , h_{t{-}1}] + b_{c}), 
\end{equation}
\begin{equation}
  c_{t} = f_t \odot{·}c_{t{-}1} + i_{t} \odot{·}c˜_{t} ,
\end{equation}
\begin{equation}
  ht = o_t \odot{·}tanh(c_{t})
\end{equation}

\subsubsection{Conditional Random Fields}

Assuming that a sequence of input words \textbf{X} = {{\textit{$x_{1}, x_{2}, x_{3} ..... x_{n} $}}} needs to be labeled a sequence of output tags \textbf{Y} = {{\textit{$y_{1}, y_{2}, y_{3} ..... y_{n} $}}}, then we can define Conditional Random Fields as discriminative sequence models that computes the posterior probability p(\textbf{Y} | \textbf{X}) directly, and thereby learns to differentiate between the possible tag sequences. The highest posterior probability is chosen as the best sequence.

\section{Experimental Setup and Results}
We train our model on a 16GB RAM, 4-core x86 CPU on the dataset prepared during the staging step. The training details are mentioned below.

\begin{figure}
\centering
\includegraphics[scale=0.4]{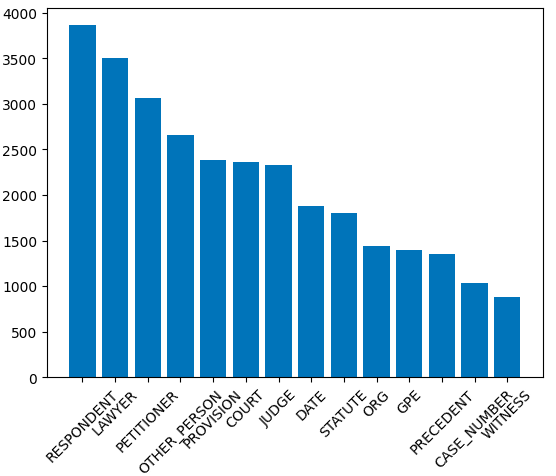}
\caption{\label{fig:The-caption}Training Class Distribution}
\end{figure}

\begin{figure}
\centering
\includegraphics[scale=0.4]{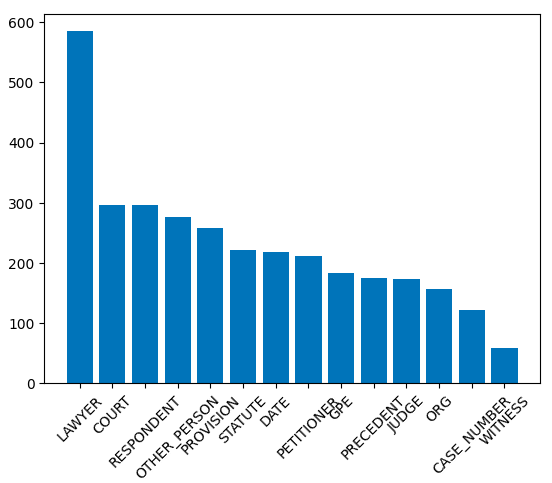}
\caption{\label{fig:The-caption}Validation Class Distribution}
\end{figure}

\begin{table}[!h]
\begin{center}
\begin{tabularx}{0.45\textwidth} { 
  | >{\centering\arraybackslash}X 
  | >{\centering\arraybackslash}X 
  | >{\centering\arraybackslash}X | }
 \hline
 \textbf{ Class} & \textbf{Training} & \textbf{Validation}\\
 \hline
 Court & 2367 & 296 \\
 \hline
 Petitioner & 3067 & 211 \\
 \hline
 Respondent & 3862 & 296 \\
 \hline
 Lawyer  & 3503 & 585 \\
 \hline
 Judge  & 2324 & 174 \\
 \hline
 Org  & 1441 & 157 \\
 \hline
 Other  & 2653 & 276 \\
 \hline
 Witness  & 881 & 58 \\
 \hline
 GPE  & 1398 & 183 \\
 \hline
 Statute  & 1804 & 222 \\
 \hline
 Date  & 1880 & 218 \\
 \hline
 Provision  & 2384 & 258 \\
 \hline
 Precedent  & 1350 & 175 \\
 \hline
 CaseNumber  & 1038 & 121 \\
 \hline
\end{tabularx}
\caption{\label{demo-table}Class Distribution}
\end{center}
\end{table}

\subsection{Stacked Embeddings}
As many sequence labeling models often combine different types of embeddings by concatenating each embedding vector to form the final word vectors. We similarly experiment with different stacked embeddings.
We add classic word embeddings such as Glove which can yield greater latent word-level semantics.

\subsection{Training}
The dataset consists of 9896 labeled training samples of the legal documents. We also split the dataset into validation and test sets to observe the F1 scores during training. Table 1. lists the distribution of classes in each of the sets. The dev and test data label distribution are also similar to that of training data. Table 2. summarizes the hyper-parameters that were selected for the best performing model. After obtaining the optimal values for the hyper-parameters, validation set is combined with the training set and the model is trained again to evaluate the final performance of the model.
\\
We record F1 scores and accuracy of the model across the validation datasets on every epoch. We adopt early stopping of training by checking the validation accuracy scores, so to avoid over-fitting on the training set.
\subsection{Analysis of Results}
\begin{figure}
\centering
\includegraphics[scale=0.2]{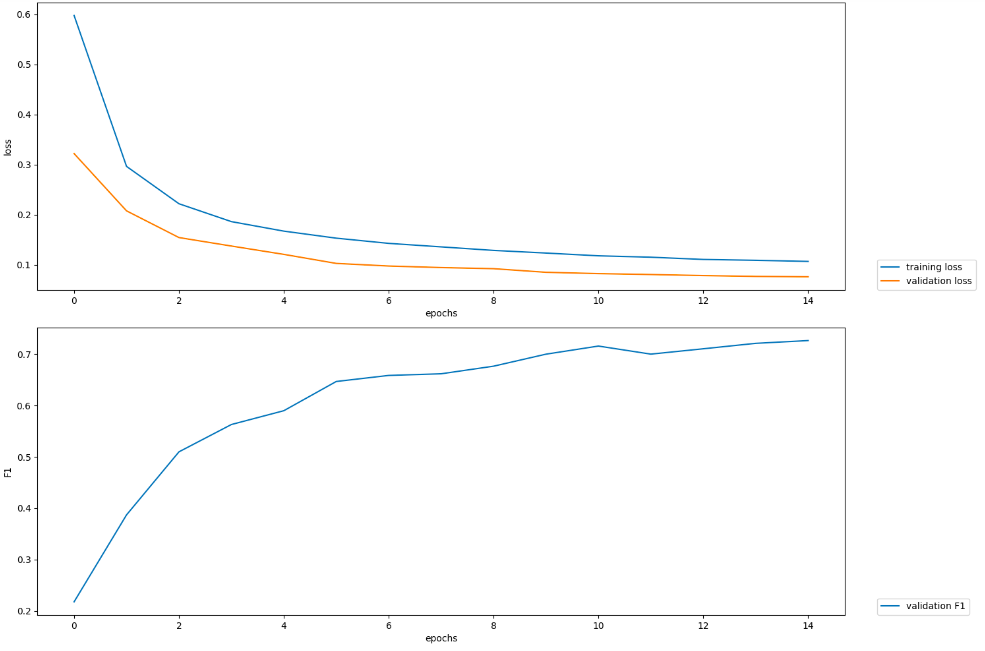}
\caption{\label{fig:The-caption} Training Loss }
\end{figure}
Our experimental results are summarized in Table 3. We find that this approach achieves 72\% F1-scores in the legal entity labeling task and that the proposed contextual string embeddings for the model is indeed  useful for sequence labeling.
\\
In figure 4. we plot the training and the validation loss with respect to epochs trained. As we observe the rise in validation loss, we save it as the best possible generalized model and report the scores on it.

\begin{table}[!h]
\begin{center}
\begin{tabularx}{0.4\textwidth} { 
  | >{\centering\arraybackslash}X 
  | >{\centering\arraybackslash}X 
  | >{\centering\arraybackslash}X | }
 \hline
 \textbf{ Parameter} & \textbf{Value}\\
 \hline
 Epochs & 50  \\
 \hline
 Learning Rate & 0.1  \\
 \hline
 Batch Size & 32  \\
 \hline
 Optimizer & SGD \\
 \hline
 Glove Dimension & 100 \\
 \hline
 Word dropout & 0.5 \\
 \hline
 LSTM Hidden Layer & 256 \\
 \hline
\end{tabularx}
\caption{\label{demo-table}Parameter Selection}
\end{center}
\end{table}

\begin{table}[!h]
\begin{center}
\begin{tabularx}{0.4\textwidth} { 
  | >{\centering\arraybackslash}X 
  | >{\centering\arraybackslash}X 
 | }
 \hline
 \textbf{ Metric } & \textbf{ Value }\\
 \hline
 Micro F1 Score & 0.724  \\
 \hline
 Weighted F1 Score & 0.762  \\
 \hline
 Macro Avg & 0.632  \\
 \hline
\end{tabularx}
\caption{\label{demo-table}Results}
\end{center}
\end{table}

\begin{table}[!h]
\begin{center}
\begin{tabularx}{0.45\textwidth} { 
  | >{\centering\arraybackslash}X 
  | >{\centering\arraybackslash}X 
  | >{\centering\arraybackslash}X 
  | >{\centering\arraybackslash}X| }
 \hline
 \textbf{ Class} & \textbf{Precision} & \textbf{Recall} & \textbf{F1-score}\\
 \hline
 Court & 0.84 & 0.86 & 0.85 \\
 \hline
 Petitioner & 0.65 & 0.24 & 0.35 \\
 \hline
 Resp & 0.33 & 0.7 & 0.12 \\
 \hline
 Judge & 0.72 & 0.60 & 0.66 \\
 \hline
 Org  & 0.56 & 0.56 & 0.56 \\
 \hline
 Other & 0.57 & 0.59 & 0.58 \\
 \hline
 Witness & 0.57 & 0.54 & 0.55 \\
 \hline
 GPE  & 0.72 & 0.65 & 0.69 \\
 \hline
 Statute & 0.82 & 0.88 & 0.85 \\
 \hline
 Date  & 0.90 & 0.85 & 0.87 \\
 \hline
 Provision & 0.85 & 0.87 & 0.86 \\
 \hline
 Precedent & 0.64 & 0.61 & 0.63 \\
 \hline
 Case & 0.58 & 0.63 & 0.61 \\
 \hline
\end{tabularx}
\caption{\label{demo-table}Results by Class}
\end{center}
\end{table}

\section{Conclusion}
In this paper, we developed a statistical based Named Entity Recognition model for labeling legal documents for the LegalNER task. We constructed our model using two LSTM layers in both directions to create a context vector for each token and used a CRF layer to find the best label sequence. We also incorporated the contextual string embedding as the input to LSTM layer, which has proved effective to vectorize polysemous tokens. We also produce an IOB formatted legal dataset which was used during the training stages of the model. We show that the system produces results with 75\% F1-scores with respect to legal NER. This is an important preprocessing step for many of NLP tasks ranging from Chatbots, Information Extraction and Entity Linking. We believe this can lead to wider adoption of Natural Language techniques in legal domains.

\bibliography{acl2023}
\bibliographystyle{acl_natbib}

\end{document}